\documentclass[letterpaper, 10 pt, journal, twoside]{ieeetran}

\usepackage{amsmath}
\usepackage{amsfonts}
\usepackage{fixmath}
\usepackage{graphicx}
\usepackage{bm}

\IEEEoverridecommandlockouts
\newtheorem{remark}{Remark}

\begin{document}

\setcounter{topnumber}{3} \setcounter{bottomnumber}{2} \renewcommand{%
\textfraction}{0.00001}

\renewcommand {\floatpagefraction}{0.990} \renewcommand{\textfraction}{0.001}
\renewcommand{\topfraction}{0.99} \renewcommand{\bottomfraction}{0.99} %
\renewcommand{\floatpagefraction}{0.99} \setcounter{totalnumber}{4}

\title{\LARGE \bf
An $O(n)$-Algorithm for the Higher-Order Kinematics and Inverse Dynamics of Serial Manipulators using Spatial Representation of Twists
}

\author{A. M\"{u}ller$^{1}$
\thanks{Manuscript received: August 4, 2020; Revised November 4, 2020; Accepted November 11, 2020.}
\thanks{This paper was recommended for publication by Editor Dezhen Song upon evaluation of the Associate Editor and Reviewers' comments.
This work was supported by the LCM-K2 Center within the framework of the Austrian COMET-K2 program.}
\thanks{$^{1}$First Author is with Institute of Robotics, Johannes Kepler University Linz, Austria
        {\tt\footnotesize a.mueller@jku.at}}
\thanks{Digital Object Identifier (DOI): see top of this page.}
}

\maketitle

\begin{abstract} 
Optimal control in general, and flatness-based control in particular, of
robotic arms necessitate to compute the first and second time derivatives of
the joint torques/forces required to achieve a desired motion. In view of
the required computational efficiency, recursive $O\left( n\right) $%
-algorithms were proposed to this end. Aiming at compact yet efficient
formulations, a Lie group formulation was recently proposed, making use of
body-fixed and hybrid representation of twists and wrenches. In this paper a
formulation is introduced using the spatial representation. The second-order
inverse dynamics algorithm is accompanied by a fourth-order forward and
inverse kinematics algorithm. An advantage of all Lie group formulations is
that they can be parameterized in terms of vectorial quantities that are
readily available. The method is demonstrated for the 7 DOF Franka Emika
Panda robot.
\end{abstract} 


\markboth{IEEE Robotics and Automation Letters. Preprint Version. Accepted December, 2020}
{M{\"u}ller: $O(n)$-Algorithm for the Higher-Order Kinematics and Inverse Dynamics of Serial Manipulators} 

\section{Introduction}

\IEEEPARstart{V}{arious} applications of robots need the time derivatives of the inverse
dynamics solution $\mathbf{Q}\left( t\right) $ for a prescribed motion $%
\mathbf{q}\left( t\right) $, i.e. the time derivatives of the equations of
motion (EOM) of the robotic arm%
\begin{equation}
\mathbf{Q}=\mathbf{M}\left( \mathbf{q}\right) \ddot{\mathbf{q}}+\mathbf{C}%
\left( \dot{\mathbf{q}},\mathbf{q}\right) \dot{\mathbf{q}}+\mathbf{Q}^{%
\mathrm{grav}%
}\left( \mathbf{q}\right) +\mathbf{Q}^{%
\mathrm{app}%
}\left( t\right)  \label{EOM}
\end{equation}%
where the vector of generalized coordinates $\mathbf{q}=\left( q_{1},\ldots
,q_{n}\right) ^{T}$ comprises the joint variables, and the vector of
generalized forces $\mathbf{Q}=\left( Q_{1},\ldots ,Q_{n}\right) ^{T}$
contains the joint torques/forces. $\mathbf{M}$ and $\mathbf{C}$ is the
generalized mass and Coriolis matrix, respectively, $\mathbf{Q}_{\mathrm{grav%
}}$ represents generalized gravity forces, and $\mathbf{Q}_{\mathrm{app}}$
is due to end-effector (EE) loads and general external loads.

\begin{figure}[tb]
\centering\includegraphics[width=0.75
\linewidth]{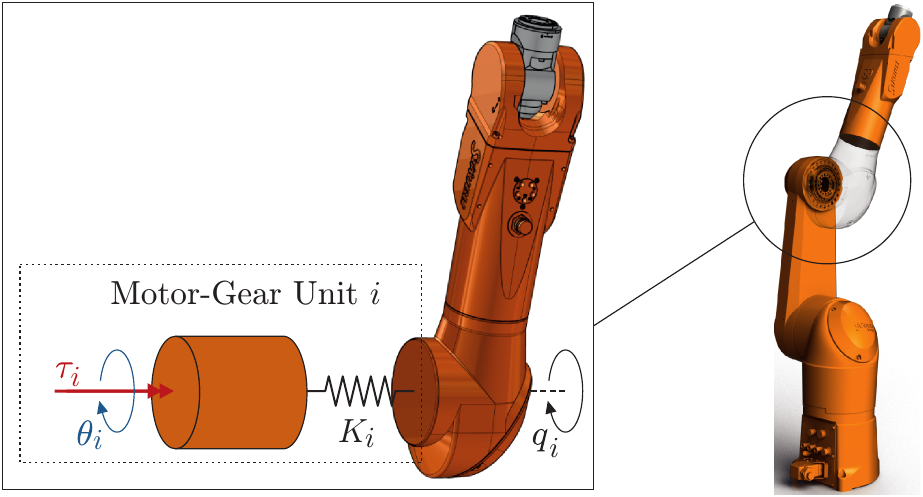}
\caption{Industrial robot regarded as a manipulator equipped with SEA}
\label{figRobotSEA}
\end{figure}

A situation where the first and second time derivatives of (\ref{EOM}) are
needed is the flatness-based control of robots equipped with serial elastic
actuators (SEA) and variable stiffness actuators (VSA) \cite%
{deLuca1998,GattringerMUBO2014,PalliMelchiorriDeLuca2008}. SEA and VSA are
in particular useful for the design of exoskeletons and safe humanoid
robots, and generally in context of human-robot interaction. Under the
assumptions stated in \cite{Spong1987}, the dynamic model of an $n$ DOF
robot actuated by SEA (Fig. \ref{figRobotSEA}) can be written as \cite%
{deLuca1998}%
\begin{align}
\mathbf{M}\left( \mathbf{q}\right) \ddot{\mathbf{q}}+\mathbf{C}\left( \dot{%
\mathbf{q}},\mathbf{q}\right) \dot{\mathbf{q}}+\mathbf{Q}^{%
\mathrm{grav}%
}\left( \mathbf{q}\right) +\mathbf{Q}^{%
\mathrm{app}%
}\left( t\right) & =\mathbf{K}_{\text{m}}\left( \bm{\theta }-\mathbf{q}%
\right)  \label{EOM1} \\
\mathbf{M}_{\text{m}}\left( \mathbf{q}\right) 
\ddot{\bm{\theta }}%
+\mathbf{K}_{\text{m}}\left( 
\bm{\theta }%
-\mathbf{q}\right) & =%
\bm{\tau }.  \label{EOM2}
\end{align}%
The additional second set of equations (\ref{EOM2}) governs the motor/gear
dynamics, where $%
\bm{\theta }%
=\left( \theta _{1},\ldots \theta _{n}\right) ^{T}$ are the
(rotation/translation) variables describing the motion of the motor/gear
units. The mass matrix $\mathbf{M}_{\text{m}}=\mathrm{diag}(M_{\mathrm{m}%
,1},\ldots ,M_{\mathrm{m},n})$ is composed of the reduced motor/gear
inertias $M_{\mathrm{m},i}$, and $\bm{\tau}$ is the vector of motor
torques/forces collocated to $\bm{\theta}$. The elastic coupling of both
systems is achieved by the gear stiffness encoded in the stiffness matrix $%
\mathbf{K}_{\text{m}}=\mathrm{diag}(K_{1},\ldots ,K_{n})$ containing the
lumped stiffness parameters $K_{i}$ of the motor/gear unit at joint $i$. The
joint torques/forces acting at the robot are now determined by the elastic
coupling of both systems as $\mathbf{Q}:=\mathbf{K}_{\text{m}}\left( 
\bm{\theta }%
-\mathbf{q}\right) $, and flatness-based control of robots equipped with
SEA/VSA involves their first and second time derivatives.

Another application where the derivatives of the drive torques/forces are
needed is the time-optimal control of standard industrial manipulators \cite%
{ReiterTII2018}. Even if the optimal control problem accounts for the limits
on speed, acceleration, jerk, and drive torques/forces, the trajectories may
still violate the limits on the rate of drive torques/forces, i.e. the
limits on the time derivatives of the motor current set by the inverter.
This requirement on the higher-order continuity of $\mathbf{Q}$ is often
overlooked, merely because time optimal motions are not yet widely applied
in practice, but it is gaining importance.%

Aiming at maximal computational efficiency, recursive $O\left( n\right) $%
-algorithms for the evaluation of the time derivatives were reported in \cite%
{BuondonnaDeLuca2015,BuondonnaDeLuca2016,Guarino2006,Guarino2009} building
upon classical Newton-Euler (NE) formulations to evaluate the EOM, where
angular and translational motions are treated separately. Using the
so-called 'spatial vector' formulation, recursive algorithms were presented 
\cite{Featherstone1983,Featherstone2008}, where angular and translational
velocities are collectively treated as rigid body twists written as
6-vectors. Observing that twists are screws, forming the Lie algebra of the
rigid body motion group $SE\left( 3\right) $, these algorithms were
reformulated in \cite{ParkBobrowPloen1995,LynchPark2017}. Such Lie group
algorithms are coordinate invariant, i.e. applicable regardless of the frame
in which twists and wrenches are represented. It is important to remember
that the 'representation' of a screw does not only regard the frame in which
the vectors are resolved but also the point relative to which they are
measured. From a computational point of view, the crucial point is that the
number of necessary operations of the frame transformation of a screw
depends on its representation. Two representations are mostly used: \emph{%
body-fixed} and \emph{spatial} representation \cite%
{Murray,MUBOScrews2,LynchPark2017} (Sec. \ref{secVel} \& Appendix \ref%
{secBodyFixedAlg}). A central feature of the Lie group (and hence of the
spatial vector) formulation is the compactness of the involved expressions.
A recursive algorithm for computation of the second time derivative of (\ref%
{EOM}) was presented in \cite{ICRA2017} using body-fixed representations.%

In the following, an $O\left( n\right) $-algorithm for the second-order
inverse dynamics is introduced, which rests on the recursive Lie group
formulation of the kinematics and dynamics using the spatial representation.
The algorithm comprises two recursions. Firstly, the forward kinematics run
distributes the 4th-order state $(\mathbf{q},\dot{\mathbf{q}},\ddot{\mathbf{q%
}},\dddot{\mathbf{q}},\ddot{\ddot{\mathbf{q}}})$ among the bodies, starting
from the ground. Secondly, the inverse dynamics run recursively evaluates
the (spatial) momentum screws of the bodies and their time derivatives, and
propagates them starting from terminal body. A salient feature of the Lie
group formulation is that it admits model description in terms of readily
available data without compromising computational efficiency. This paper
shall complement the Lie group formulations of the recursive inverse
dynamics algorithms in body-fixed and hybrid representation \cite{ICRA2017},
and thus provide the starting point for research addressing the
computationally most efficient as well as (regarding the parameterization
and simplicity) the most user-friendly approach.

The paper is organized as follows. Sec. \ref{secKin} recalls the Lie group
modeling of robot kinematics. Sec. \ref{secFwKin} presents the 4th-order
forward kinematics in spatial representation of twists. The recursive
4th-order forward kinematics is combined with the inverse kinematics
solution in Sec. \ref{secInvFwKin}. The inverse dynamics algorithm is
derived in Sec. \ref{secInvDynSpatial}, and its features are briefly
discussed. Simulation results are presented in Sec. \ref{secIPanda} for the
Franka Emika Panda robot, along with a preliminary comparison of the spatial
and body-fixed algorithm. The paper closes with suggestions for future work
in Sec. \ref{secConclusion}.%

\section{Manipulator Kinematics%
\label{secKin}%
}

\subsection{Manipulator configuration via the POE}

A stationary robotic arm is modeled as a simple kinematic chain comprising $%
n $ rigid bodies connected to the ground by $n$ 1-DOF helical (revolute,
screw, prismatic) joints. The pose of body $i$ is represented by a
homogenous transformation matrix $\mathbf{C}_{i}$ in (\ref{Ci}). More
precisely, $\mathbf{C}_{i}$ describes the transformation from a body-fixed
reference frame (RFR) $\mathcal{F}_{i}$, which is kinematically representing
body $i$, to a world-fixed inertia frame (IFR).

Denote with $\mathbf{q}\in \mathbb{V}^{n}$ the vector of joint variables $%
q_{i},i=1,\ldots ,n$, with ${\mathbb{V}}^{n}={\mathbb{T}}^{n_{r}}\times {%
\mathbb{R}}^{n_{p}}$, where $n_{r}$ and $n_{p}$ is the number of
revolute/screw and prismatic joints, respectively. For a given vector of
joint variables $\mathbf{q}\in {\mathbb{V}}^{n}$, the configuration of link $%
i$ can be expressed as $\mathbf{C}_{i}\left( \mathbf{q}\right) =f_{i}\left( 
\mathbf{q}\right) \mathbf{A}_{i}$ using the product of exponentials (POE) in
the form \cite{MUBOScrews2,Mueller-MMT2019}%
\begin{equation}
f_{i}\left( \mathbf{q}\right) =\exp \left( \mathbf{Y}_{1}q_{1}\right) \exp
\left( \mathbf{Y}_{2}q_{2}\right) \cdot \ldots \cdot \exp \left( \mathbf{Y}%
_{i}q_{i}\right) ,  \label{fi}
\end{equation}%
which was introduced in \cite{Brockett1984} and is now a central element in
modern robotics \cite{LynchPark2017,Murray}. The POE (\ref{fi}) describes
the relative configuration of the RFR $\mathcal{F}_{i}$ at body $i$ relative
to its location at $\mathbf{q}=\mathbf{0}$, while $\mathbf{A}_{i}:=%
\mathbf{C}_{i}\left( \mathbf{0}\right) \in SE\left( 3\right)$ is the
absolute reference configuration of the body-fixed frame w.r.t. to the IFR
in this zero-reference configuration. In the above POE, the joint kinematics
is encoded in the screw coordinate vectors $\mathbf{Y}_{j}\in {\mathbb{R}}%
^{6}$ represented in the IFR. 
The screw coordinates for a revolute joint $j$ are $\mathbf{Y}_{j}=\left( 
\mathbf{e}_{j},\mathbf{y}_{j}\times \mathbf{e}_{j}\right) ^{T}$, where $%
\mathbf{e}_{j}$ is a unit vector along the screw axis, and $\mathbf{y}_{j}$
is the position vector to a point on the revolute joint axis measured and
resolved in the IFR. The screw coordinate vector for a prismatic joint is $%
\mathbf{Y}_{j}=\left( \mathbf{e}_{j},\mathbf{0}\right) ^{T}$, while for a
screw joint with pitch $h_{j}$, this is $\mathbf{Y}_{j}=\left( \mathbf{e}%
_{j},\mathbf{y}_{j}\times \mathbf{e}_{j}+h_{j}\mathbf{e}_{j}\right) ^{T}$
These screw coordinates allow using geometric information and vectorial
quantities, deduced from an arbitrary IFR.%

\begin{remark}
The Denavit-Hartenberg (DH) convention is frequently used to describe the
manipulator geometry. The DH-parameters for joint $j$ are denoted with $%
\alpha _{j},a_{j},d_{j},\vartheta _{j}$ \cite{UickerBook}. With the
DH-convention, the absolute reference configuration is determined as%
\begin{equation}
\mathbf{A}_{i}=\mathbf{A}_{i-1}\mathrm{Rot}\left( \vartheta _{i}\right) 
\mathrm{Trans}\left( d_{i}\right) \mathrm{Trans}\left( a_{i}\right) \mathrm{%
Rot}\left( \alpha _{i}\right) 
\end{equation}%
with $\mathbf{A}_{0}=\mathbf{I}$, and with DH parameter values according to
the reference configuration. The homogenous transformation matrices $\mathrm{%
Rot}\left( \varphi \right) $ and $\mathrm{Trans}\left( x\right) $ describe a
pure rotation and translation, respectively, about/along the joint axis
according to the DH-convention \cite{LynchPark2017}.
\end{remark}

\subsection{Manipulator velocities and Jacobian in spatial representation%
\label{secVel}%
}

The twist $\mathbf{V}_{i}^{\text{$\mathrm{s}$}}=\left( \bm{\omega}_{i}^{%
\text{$\mathrm{s}$}},\mathbf{v}_{i}^{\text{$\mathrm{s}$}}\right) ^{T}$ of
link $i$ in spatial representation comprises the angular velocity $\mathbf{%
\omega }_{i}^{\text{s}}$ of $\mathcal{F}_{i}$ relative to $\mathcal{F}_{0}$,
and the translational velocity $\mathbf{v}_{i}^{\text{s}}$ of the body-fixed
point momentarily coinciding with the origin of $\mathcal{F}_{0}$, where
both vectors are resolved in $\mathcal{F}_{0}$. It is determined recursively
by \cite{LynchPark2017,MUBOScrews1}%
\begin{equation}
\mathbf{V}_{i}^{\text{$\mathrm{s}$}}=\mathbf{V}_{i-1}^{\text{$\mathrm{s}$}}+%
\mathbf{S}_{i}\dot{q}_{i},i=1,\ldots ,n  \label{Vsirec}
\end{equation}%
with $\mathbf{V}_{0}^{\text{\textrm{s}}}=\mathbf{0}$, where $\mathbf{S}%
_{i}\in {\mathbb{R}}^{6}$ is the instantaneous joint screw coordinate vector
of joint $i$. The latter is determined from the joint screw coordinates $%
\mathbf{Y}_{j}$ in the reference configuration by the frame transformation
according to the motion of body $j$, according to (\ref{fi}), as%
\begin{equation}
\mathbf{S}_{j}%
\hspace{-0.5ex}%
\left( \mathbf{q}\right) =\mathbf{Ad}_{f_{j}\left( \mathbf{q}\right) }%
\mathbf{Y}_{j},\ j\leq i  \label{Si}
\end{equation}%
where the adjoint operator $\mathbf{Ad}_{\mathbf{C}}$ in (\ref{Ad})
describes the transformation of screw coordinates according to the frame
transform $\mathbf{C}\in SE\left( 3\right) $ \cite{Murray,Selig}%
. The EE is attached to the terminal body $n$, and w.l.o.g. the twist of the
terminal body is identified with the EE-twist, $\mathbf{V}_{\mathrm{E}}^{%
\text{\textrm{s}}}:=\mathbf{V}_{n}^{\text{\textrm{s}}}$. For the EE/terminal
body $n$, the recursive summation (\ref{Vsirec}) can be summarized as%
\begin{equation}
\mathbf{V}_{\mathrm{E}}^{\text{$\mathrm{s}$}}=\mathbf{J}\dot{%
\mathbf{q}}\text{\ \ \ with \ \ }\mathbf{J}\left( \mathbf{q}%
\right) :=%
\Big%
(\mathbf{S}_{1}%
\hspace{-0.5ex}%
\left( \mathbf{q}\right) 
\Big%
|\cdots 
\Big%
|\mathbf{S}_{n}%
\hspace{-0.5ex}%
\left( \mathbf{q}\right) 
\Big%
)  \label{Ve}
\end{equation}%
where $\mathbf{J}$ is the \emph{spatial Jacobian} of the robotic
arm. 
This is the geometric Jacobian \cite{SicilianoSciaviccoVillaniOriolo2009} of
the robotic arm that determines the EE-twist in spatial representation (with
the joint screws $\mathbf{S}_{j}$ in spatial representation). The geometric
Jacobian that determines the EE-twist in body-fixed representation is called
the \emph{body-fixed Jacobian} \cite{MUBOScrews2}.%

In contrast to the recursive relation in terms of body-fixed twists 
(see (\ref{Vbrec}) in Appendix \ref{secBodyFixedAlg}), 
the relation (\ref{Vsirec}) does not involve frame transformations of
twists. However, the screw coordinates $\mathbf{S}_{j}$ are not constant and
must be computed with (\ref{Si}), so that both formulations involve a
similar number of operations.

\section{Fourth-Order Forward Kinematics 
\label{secFwKin}%
}

The recursion (\ref{Vsirec}) solves the first-order forward kinematics
problem of the \emph{robot mechanism}, which consists in determining the
state of all bodies of the robot mechanism for given state $\left( \mathbf{q}%
,\dot{\mathbf{q}}\right) $, whereas (\ref{Ve}) solves the forward kinematics
problem of the \emph{robot manipulator}. Now, the second-order derivatives
of the EOM (\ref{EOM}) involve the 4th-order time derivatives of $\mathbf{q}%
\left( t\right) $, and a recursive evaluation requires solution of the
4th-order forward kinematics problem, which is to determine the velocity,
acceleration, jerk, and jounce of all bodies for given $\mathbf{q},\dot{%
\mathbf{q}},\ddot{\mathbf{q}},\dddot{\mathbf{q}},\ddot{\ddot{\mathbf{q}}}$.
This needs time derivatives of $\mathbf{S}_{i}$. 
The latter is given as $\dot{\mathbf{S}}_{i}=\mathbf{ad}_{\mathbf{V}^{%
\mathrm{s}}}\mathbf{S}_{i}$, where $\mathbf{ad}$ is the 'screw product'
matrix in (\ref{adse3}). 
Repeated application of this, along with (\ref{Vsirec}), 
and using the $\mathbf{ad}$ matrix in (\ref{adse3}), yields \cite{Mueller-MMT2019}%
\begin{align}
\dot{\mathbf{S}}_{i}& =\mathbf{ad}_{\mathbf{V}_{i}^{\text{$\mathrm{s}$}}}%
\mathbf{S}_{i},\ \ddot{\mathbf{S}}_{i}=%
\big%
(\mathbf{ad}_{\dot{\mathbf{V}}_{i}^{\text{$\mathrm{s}$}}}+\mathbf{ad}_{%
\mathbf{V}_{i}^{\text{$\mathrm{s}$}}}^{2}%
\big%
)\mathbf{S}_{i}  \label{Sdot} \\
\dddot{\mathbf{S}}_{i}& =%
\big%
(\mathbf{ad}_{\ddot{\mathbf{V}}_{i}^{\text{$\mathrm{s}$}}}+2\mathbf{ad}_{%
\dot{\mathbf{V}}_{i}^{\text{$\mathrm{s}$}}}\mathbf{ad}_{\mathbf{V}_{i}^{%
\text{$\mathrm{s}$}}}+\mathbf{ad}_{\mathbf{V}_{i}^{\text{$\mathrm{s}$}}}%
\mathbf{ad}_{\dot{\mathbf{V}}_{i}^{\text{$\mathrm{s}$}}}+\mathbf{ad}_{%
\mathbf{V}_{i}^{\text{$\mathrm{s}$}}}^{3}%
\big%
)\mathbf{S}_{i}.  \notag
\end{align}%
This gives rise to the recursive $O\left( n\right) $-algorithm 1 for the
4th-order forward kinematics for the robot mechanism where for the ground
body ($i=0$) $\mathbf{V}_{0}^{\text{\textrm{s}}}=\ddot{\mathbf{V}}_{0}^{%
\text{\textrm{s}}}=\dddot{\mathbf{V}}_{0}^{\text{\textrm{s}}}=\mathbf{0}$,
and $f_{0}=\mathbf{I}$. In order to account for gravity, the ground
acceleration is set to $\dot{\mathbf{V}}_{0}^{\text{\textrm{s}}}=\mathbf{G}%
_{0}^{\mathrm{s}}=\left( \mathbf{0},-\mathbf{g}\right)^T$ with the
gravitational acceleration vector $\mathbf{g}\in {\mathbb{R}}^{3}$
represented in the IFR. The instantaneous screw coordinates $\mathbf{S}_{i}$
and their time derivatives will be reused in the inverse dynamics recursion. %
\begin{figure}[ht]
\begin{tabular}{l}
\textbf{Algorithm 1: }%
\textbf{4th-Order Forward Kinematics Recursion}%
\\ \hline
\textbf{Input:} $\mathbf{q},\dot{\mathbf{q}},\ddot{\mathbf{q}},\dddot{%
\mathbf{q}},\ddot{\ddot{\mathbf{q}}}$, and $\mathbf{Y}_{i},i=1,\ldots ,n$ \\ 
For $i=1,\ldots ,n$ do \\ 
\begin{tabular}{rll}
$f_{i}%
\hspace{-2ex}%
$ & $=%
\hspace{-2ex}%
$ & $f_{i-1}\exp (\mathbf{Y}_{i}q_{i})$ \ \ \ \ \ \ \ \ \ Position kinematics
\\ 
$\mathbf{C}_{i}%
\hspace{-2ex}%
$ & $=%
\hspace{-2ex}%
$ & $f_{i}\mathbf{A}_{i}$ \\ 
$\mathbf{S}_{i}%
\hspace{-2ex}%
$ & $=%
\hspace{-2ex}%
$ & $\mathbf{Ad}_{f_{i}\left( \mathbf{q}\right) }\mathbf{Y}_{i}\ \ \ \ \ \ \
\ \ \ \ \ \ \ \ $Higher-order kinematics \\ 
$\mathbf{V}_{i}^{\text{\textrm{s}}}%
\hspace{-2ex}%
$ & $=%
\hspace{-2ex}%
$ & $\mathbf{V}_{i-1}^{\text{\textrm{s}}}+\mathbf{S}_{i}\dot{q}_{i}$ \\ 
$\dot{\mathbf{S}}_{i}%
\hspace{-2ex}%
$ & $=%
\hspace{-2ex}%
$ & $\mathbf{ad}_{\mathbf{V}_{i}^{\text{\textrm{s}}}}\mathbf{S}_{i}$ \\ 
$\dot{\mathbf{V}}_{i}^{\text{\textrm{s}}}%
\hspace{-2ex}%
$ & $=%
\hspace{-2ex}%
$ & $\dot{\mathbf{V}}_{i-1}^{\text{\textrm{s}}}+\mathbf{S}_{i}\ddot{q}_{i}+%
\dot{\mathbf{S}}_{i}\dot{q}_{i}$ \\ 
$\ddot{\mathbf{S}}_{i}%
\hspace{-2ex}%
$ & $=%
\hspace{-2ex}%
$ & $%
\big%
(\mathbf{ad}_{\dot{\mathbf{V}}_{i}^{\text{\textrm{s}}}}+\mathbf{ad}_{\mathbf{%
V}_{i}^{\text{\textrm{s}}}}^{2}%
\big%
)\mathbf{S}_{i}$ \\ 
$\ddot{\mathbf{V}}_{i}^{\text{\textrm{s}}}%
\hspace{-2ex}%
$ & $=%
\hspace{-2ex}%
$ & $\ddot{\mathbf{V}}_{i-1}^{\text{\textrm{s}}}+\mathbf{S}_{i}\dddot{q}%
_{i}+2\dot{\mathbf{S}}_{i}\ddot{q}_{i}+\ddot{\mathbf{S}}_{i}\dot{q}_{i}$ \\ 
$\dddot{\mathbf{S}}_{i}%
\hspace{-2ex}%
$ & $=%
\hspace{-2ex}%
$ & $%
\big%
(\mathbf{ad}_{\ddot{\mathbf{V}}_{i}^{\text{\textrm{s}}}}+2\mathbf{ad}_{\dot{%
\mathbf{V}}_{i}^{\text{\textrm{s}}}}\mathbf{ad}_{\mathbf{V}_{i}^{\text{%
\textrm{s}}}}+\mathbf{ad}_{\mathbf{V}_{i}^{\text{\textrm{s}}}}\mathbf{ad}_{%
\dot{\mathbf{V}}_{i}^{\text{\textrm{s}}}}+\mathbf{ad}_{\mathbf{V}_{i}^{\text{%
\textrm{s}}}}^{3}%
\big%
)\mathbf{S}_{i}$ \\ 
$\dddot{\mathbf{V}}_{i}^{\text{\textrm{s}}}%
\hspace{-2ex}%
$ & $=%
\hspace{-2ex}%
$ & $\dddot{\mathbf{V}}_{i-1}^{\text{\textrm{s}}}+\mathbf{S}_{i}\ddot{\ddot{q%
}}_{i}+3\dot{\mathbf{S}}_{i}\dddot{q}_{i}+3\ddot{\mathbf{S}}_{i}\ddot{q}_{i}+%
\dddot{\mathbf{S}}_{i}\dot{q}_{i}$%
\end{tabular}
\\ 
End \\ 
\textbf{Output:} $\mathbf{V}_{i}^{\text{\textrm{s}}},\dot{\mathbf{V}}_{i}^{%
\text{\textrm{s}}},\ddot{\mathbf{V}}_{i}^{\text{\textrm{s}}},\dddot{\mathbf{V%
}}_{i}^{\text{\textrm{s}}}$ and $\mathbf{S}_{i},\dot{\mathbf{S}}_{i},\ddot{%
\mathbf{S}}_{i},\dddot{\mathbf{S}}_{i}$ \\ \hline
\end{tabular}%
\vspace{-3ex}
\end{figure}

\section{Combined Fourth-Order Inverse and Forward Kinematics Recursion%
\label{secInvFwKin}%
}

The above forward kinematics run assumes that the trajectory $\mathbf{q}%
\left( t\right) $ and its derivatives are known. If the motion is not
determined up-front in terms of joint variables (e.g. as a result of optimal
motion planning in joint space), it will have to be determined from a
prescribed EE-motion, which necessitates solving the \emph{inverse
kinematics problem of the manipulator}. 
The higher-order inverse kinematics is addressed in the following for
kinematically non-redundant manipulators, where $\mathbf{J}$ is a $n\times n$
matrix. The geometric inverse kinematics problem can be solved numerically
using closed-loop inverse kinematics (CLIK) algorithms \cite%
{ColomeTorras2015} or with a Newton iteration scheme using the inverse of $%
\mathbf{J}$.

\begin{remark}
The inverse kinematics solution can also be used for kinematically redundant
robots when the joint space decomposition is used, which was proposed in 
\cite{Wampler1987}, and used for trajectory planning in \cite%
{Galicki2000,MaWatanabe2004}. Details of the higher-order inverse kinematics
in closed form can be found in \cite{ReiterTII2018}.
\end{remark}

The solution to the 1st-order (velocity) inverse kinematics problem of a
non-redundant robot follows from (\ref{Ve}) as $\dot{\mathbf{q}}=\mathbf{J}_{%
\mathrm{E}}^{-1}\mathbf{V}_{\mathrm{E}}^{\text{s}}$, presumed that it is not
in a forward kinematic singularity. The EE-accelaration can be expressed as $%
\dot{\mathbf{V}}_{\mathrm{E}}^{\text{s}}=\mathbf{J}\ddot{\mathbf{q}}%
+\sum_{i\leq n}\dot{\mathbf{S}}_{i}\dot{q}_{i}$, and the solution for
the 2nd-order (acceleration) inverse kinematics problem is $\ddot{\mathbf{q}}%
=\mathbf{J}^{-1}%
\Big%
(\dot{\mathbf{V}}_{\mathrm{E}}^{\text{s}}-\sum_{i\leq n}\dot{\mathbf{S}}_{i}%
\dot{q}_{i}%
\Big%
)$. The expressions for the jerk and jounce are similar. The above
expressions (\ref{Sdot}) for the derivatives of $\mathbf{S}_{i}$ could be
inserted to yield closed form solutions \cite{Mueller-MMT2019}, but it is
computationally more efficient to compute and reuse them.

The relation for $\mathbf{q}^{\left( k\right) }$ involves the first $k-1$
time derivative of the EE-twist $\mathbf{V}_{\mathrm{E}}^{\text{s}}$ and,
via (\ref{Sdot}), the derivatives of the twists $\mathbf{V}_{i}^{\text{s}}$
of up to order $k-2$ of all bodies. Thus, solving the manipulator inverse
kinematics of order $k$ presupposes the solution of the forward kinematics
of the mechanism up to order $k-1$. These steps are summarized in Algorithm
2. The Jacobian inverse $\mathbf{J}^{-1}\left( \mathbf{q}\right) $ is only
computed once in step 2.a). The terms $\dot{\mathbf{S}}_{i}\dot{q}_{i}$ from
3a) can be reused in 3b), $\ddot{\mathbf{S}}_{i}\dot{q}_{i},\dot{\mathbf{S}}%
_{i}\ddot{q}_{i}$ from 4a) in 4b), and $\dot{\mathbf{S}}_{i}\dddot{q}_{i},%
\ddot{\mathbf{S}}_{i}\ddot{q}_{i},\dddot{\mathbf{S}}_{i}\dot{q}_{i}$ from
5.a) in 5.b).
\begin{figure}[ht]
\begin{tabular}{l}
\textbf{Algorithm 2: } \\ 
\textbf{Combined inverse and forward kinematics recursion} \\ \hline
\textbf{Input:} $\mathbf{q},\mathbf{V}_{\mathrm{E}}^{\text{s}},\dot{\mathbf{V%
}}_{\mathrm{E}}^{\text{s}},\ddot{\mathbf{V}}_{\mathrm{E}}^{\text{s}},\dddot{%
\mathbf{V}}_{\mathrm{E}}^{\text{s}}$, and $\mathbf{Y}_{i},i=1,\ldots ,n$ \\ 
\textbf{1)} \ Configurations, joint screws (EE-Jacobian): \\ 
\hspace{9ex}%
\begin{tabular}{l}
For $i=1,\ldots ,n$ do \\ 
\hspace{2ex}%
$f_{i}=f_{i-1}\exp (\mathbf{Y}_{i}q_{i})$ \\ 
\hspace{2ex}%
$\mathbf{C}_{i}=f_{i}\mathbf{A}_{i}$ \\ 
\hspace{2ex}%
$\mathbf{S}_{i}=\mathbf{Ad}_{f_{i}\left( \mathbf{q}\right) }\mathbf{Y}_{i}$
\\ 
End%
\end{tabular}
\\ 
\textbf{2.a)} \ 1st-order inverse kinematics of manipulator: \\ 
\hspace{10ex}%
$\dot{\mathbf{q}}=\mathbf{J}^{-1}\mathbf{V}_{\mathrm{E}}^{\text{s}}$ \\ 
\textbf{2.b)} \ 1st-order forward kinematics of robot mechanism: \\ 
\hspace{9ex}%
\begin{tabular}{l}
For $i=1,\ldots ,n-1$ do \\ 
\hspace{2ex}%
$\mathbf{V}_{i}^{\text{s}}=\mathbf{V}_{i-1}^{\text{s}}+\mathbf{S}_{i}\dot{q}%
_{i}$ \\ 
\hspace{2ex}%
$~\dot{\mathbf{S}}_{i}=\mathbf{ad}_{\mathbf{V}_{i}^{\text{s}}}\mathbf{S}_{i}$
\\ 
End%
\end{tabular}
\\ 
\textbf{3.a)} \ 2nd-order inverse kinematics of manipulator: \\ 
\hspace{10ex}%
$\ddot{\mathbf{q}}=\mathbf{J}^{-1}%
\Big%
(\dot{\mathbf{V}}_{\mathrm{E}}^{\text{s}}-\sum_{i\leq n}\dot{\mathbf{S}}_{i}%
\dot{q}_{i}%
\Big%
)$ \\ 
\textbf{3.b)} \ 2nd-order forward kinematics of robot mechanism: \\ 
\hspace{9ex}%
\begin{tabular}{l}
For $i=1,\ldots ,n-1$ do \\ 
\hspace{2ex}%
\begin{tabular}{rl}
$\dot{\mathbf{V}}_{i}^{\text{s}}=%
\hspace{-2ex}%
$ & $\mathbf{V}_{i-1}^{\text{s}}+\mathbf{S}_{i}\ddot{q}_{i}+\dot{\mathbf{S}}%
_{i}\dot{q}_{i}$ \\ 
$\ddot{\mathbf{S}}_{i}=%
\hspace{-2ex}%
$ & $%
\big%
(\mathbf{ad}_{\dot{\mathbf{V}}_{i}^{\text{s}}}+\mathbf{ad}_{\mathbf{V}_{i}^{%
\text{s}}}^{2}%
\big%
)\mathbf{S}_{i}$%
\end{tabular}
\\ 
End%
\end{tabular}
\\ 
\textbf{4.a)} \ 3rd-order inverse kinematics of manipulator: \\ 
\hspace{10ex}%
$\dddot{\mathbf{q}}=\mathbf{J}^{-1}%
\Big%
(\ddot{\mathbf{V}}_{\mathrm{E}}^{\text{s}}-\sum_{i\leq n}(2\dot{\mathbf{S}}%
_{i}\ddot{q}_{i}+\ddot{\mathbf{S}}_{i}\dot{q}_{i})%
\Big%
)$ \\ 
\textbf{4.b)} \ 3rd-order forward kinematics of robot mechanism: \\ 
\hspace{1ex}%
\begin{tabular}{l}
For $i=1,\ldots ,n-1$ do \\ 
\hspace{1ex}%
\begin{tabular}{rl}
$\ddot{\mathbf{V}}_{i}^{\text{s}}=%
\hspace{-2ex}%
$ & $\ddot{\mathbf{V}}_{i-1}^{\text{s}}+\mathbf{S}_{i}\dddot{q}_{i}+2\dot{%
\mathbf{S}}_{i}\ddot{q}_{i}+\ddot{\mathbf{S}}_{i}\dot{q}_{i}$ \\ 
$\dddot{\mathbf{S}}_{i}=%
\hspace{-2ex}%
$ & $%
\big%
(\mathbf{ad}_{\ddot{\mathbf{V}}_{i}^{\text{s}}}+2\mathbf{ad}_{\dot{\mathbf{V}%
}_{i}^{\text{s}}}\mathbf{ad}_{\mathbf{V}_{i}^{\text{s}}}+\mathbf{ad}_{%
\mathbf{V}_{i}^{\text{s}}}\mathbf{ad}_{\dot{\mathbf{V}}_{i}^{\text{s}}}+%
\mathbf{ad}_{\mathbf{V}_{i}^{\text{s}}}^{3}%
\big%
)\mathbf{S}_{i}$%
\end{tabular}
\\ 
End%
\end{tabular}
\\ 
\textbf{5.a)} \ 4th-order inverse kinematics of manipulator: \\ 
$\ddot{\ddot{\mathbf{q}}}=\mathbf{J}^{-1}%
\Big%
(\dddot{\mathbf{V}}_{n}^{\text{s}}-\sum_{i\leq n}(\dot{\mathbf{S}}_{i}\dddot{%
q}_{i}+3\ddot{\mathbf{S}}_{i}\ddot{q}_{i}+\dddot{\mathbf{S}}_{i}\dot{q}_{i})%
\Big%
)$ \\ 
\hspace{4ex}
\\ 
\textbf{5.b)} \ 4th-order forward kinematics of robot mechanism: \\ 
\begin{tabular}{l}
For $i=1,\ldots ,n-1$ do \\ 
\hspace{2ex}%
$\dddot{\mathbf{V}}_{i}^{\text{s}}=\dddot{\mathbf{V}}_{i-1}^{\text{s}}+%
\mathbf{S}_{i}\ddot{\ddot{q}}_{i}+3\dot{\mathbf{S}}_{i}\dddot{q}_{i}+3\ddot{%
\mathbf{S}}_{i}\ddot{q}_{i}+\dddot{\mathbf{S}}_{i}\dot{q}_{i}$ \\ 
End%
\end{tabular}
\\ 
\textbf{Output:} $\dot{\mathbf{q}},\ddot{\mathbf{q}},\dddot{\mathbf{q}},%
\ddot{\ddot{\mathbf{q}}}$, $\dot{\mathbf{S}}_{i},\ddot{\mathbf{S}}_{i},%
\dddot{\mathbf{S}}_{i}$, $\mathbf{V}_{i}^{\text{s}},\dot{\mathbf{V}}_{i}^{%
\text{s}},\ddot{\mathbf{V}}_{i}^{\text{s}},\dddot{\mathbf{V}}_{i}^{\text{s}}$%
, $\mathbf{J}\left( \mathbf{q}\right) $ \\ \hline
\end{tabular}
\vspace{-3ex}
\end{figure}

\section{Higher-Order Inverse Dynamics%
\label{secInvDynSpatial}%
}

\subsection{Spatial NE-equations and their time derivatives}

Consider a free-floating rigid body whose pose relative to the IFR is
described by $\mathbf{C}\in SE\left( 3\right) $. Its spatial momentum
co-screw 
\begin{equation}
\mathbf{\Pi }^{\mathrm{s}}=\mathbf{M}^{\mathrm{s}}\mathbf{V}^{\mathrm{s}}\in
se^{\ast }\left( 3\right)  \label{Pis}
\end{equation}%
is determined by the \emph{spatial mass matrix} 
\begin{equation}
\mathbf{M}^{\mathrm{s}}=\left( 
\begin{array}{cc}
\mathbf{\Theta }^{\mathrm{s}} & m(\widetilde{\mathbf{d}}+\widetilde{\mathbf{r%
}}) \\ 
-m(\widetilde{\mathbf{d}}+\widetilde{\mathbf{r}})\ \ \  & m\mathbf{I}%
\end{array}%
\right) .
\end{equation}%
Therein, $\mathbf{r}\in {\mathbb{R}}^{3}$ is the position vector of the
body-fixed frame, and $\mathbf{d}\in {\mathbb{R}}^{3}$ is the vector from
the body-fixed frame to the COM, both resolved in IFR, and $\mathbf{\Theta }%
^{\mathrm{s}}$ is the inertia tensor, resolved in IFR, w.r.t. the body-fixed
point which instantaneously coincides with the IFR origin. The spatial mass
matrix is configuration dependent. The mass matrix is constant if it is
represented in the body-fixed frame $\mathcal{F}_{\mathrm{b}}$. This \emph{%
body-fixed mass matrix} has the form%
\begin{equation}
\mathbf{M}^{\text{b}}=\left( 
\begin{array}{cc}
\mathbf{\Theta }^{\text{b}} & {^{\mathrm{b}}\widetilde{\mathbf{d}}}m \\ 
-{^{\mathrm{b}}}\widetilde{\mathbf{d}}m & \mathbf{I}m%
\end{array}%
\right) .  \label{Mb}
\end{equation}%
where ${^{\mathrm{b}}\mathbf{d}}$ is the vector from the origin of the
body-fixed frame to the COM resolved in body-fixed frame, and $\mathbf{%
\Theta }^{\text{b}}$ is the constant inertia tensor w.r.t. the body-fixed
frame. The two representations are related via \cite{MUBOScrews2} 
\begin{equation}
\mathbf{M}^{\mathrm{s}}=\mathbf{Ad}_{\mathbf{C}}^{-T}\mathbf{M}^{\text{b}}%
\mathbf{Ad}_{\mathbf{C}}^{-1}.  \label{Ms}
\end{equation}%
Along with relation (\ref{AdInvdot}), this yields the time derivative 
\begin{equation}
\dot{\mathbf{M}}^{\mathrm{s}}=-\mathbf{M}^{\mathrm{s}}\mathbf{ad}_{\mathbf{V}%
^{\mathrm{s}}}-\mathbf{ad}_{\mathbf{V}^{\mathrm{s}}}^{T}\mathbf{M}^{\mathrm{s%
}}.  \label{Msdot}
\end{equation}%
The momentum balance yields the spatial representation of the Newton-Euler
(NE) equations%
\begin{equation}
\dot{\mathbf{\Pi }}^{\mathrm{s}}+\mathbf{W}^{\mathrm{s},\mathrm{app}}+%
\mathbf{W}^{\mathrm{s},\mathrm{grav}}=\mathbf{0}%
\label{equil1}
\end{equation}%
where $%
\mathbf{W}^{\mathrm{s},\mathrm{grav}}$ 
is the wrench due to gravity and $%
\mathbf{W}^{\mathrm{s}\text{,}\mathrm{app}}$ 
the wrench due to general external loads applied to the body. Inserting (\ref%
{Pis}), and using (\ref{Msdot}), yields the NE-equations in terms of the
spatial twist and acceleration%
\begin{equation}
\mathbf{M}^{\mathrm{s}}\dot{\mathbf{V}}^{\mathrm{s}}-\mathbf{ad}_{\mathbf{V}%
^{\mathrm{s}}}^{T}\mathbf{M}^{\mathrm{s}}\mathbf{V}^{\mathrm{s}}+\mathbf{W}^{%
\mathrm{s},\mathrm{app}}+\mathbf{W}^{\mathrm{s},\mathrm{grav}}=\mathbf{0}.%
\label{NESpatial}
\end{equation}

The time derivative of the NE-equations (\ref{NESpatial}) yields the rate of
the spatial wrench in the dynamic equilibrium (\ref{equil1})%
\begin{equation}
\ddot{\mathbf{\Pi }}{^{\mathrm{s}}+\dot{\mathbf{W}}}^{\mathrm{s},\mathrm{app}%
}+\dot{\mathbf{W}}^{\mathrm{s},\mathrm{grav}}=\mathbf{0}.%
\label{equil2}
\end{equation}%
Evaluation of (\ref{equil2}) necessitates the second time derivative of the
spatial momentum screw. This derivative is obtained, with the explicit
relation (\ref{Msdot}) for the time derivative of the spatial mass matrix, as

\begin{align}
\ddot{\mathbf{\Pi }}{^{\mathrm{s}}}& =\mathbf{M}^{\mathrm{s}}\ddot{\mathbf{V}%
}^{\mathrm{s}}+\dot{\mathbf{M}}^{\mathrm{s}}\dot{\mathbf{V}}^{\mathrm{s}}-%
\mathbf{ad}_{\dot{\mathbf{V}}^{\mathrm{s}}}^{T}\mathbf{\Pi }^{\mathrm{s}}-%
\mathbf{ad}_{\mathbf{V}^{\mathrm{s}}}^{T}\dot{\mathbf{\Pi }}^{\mathrm{s}} \\
& =\mathbf{M}^{\mathrm{s}}(\ddot{\mathbf{V}}^{\mathrm{s}}-\mathbf{ad}_{%
\mathbf{V}^{\mathrm{s}}}\dot{\mathbf{V}}^{\mathrm{s}})-\mathbf{ad}_{\dot{%
\mathbf{V}}^{\mathrm{s}}}^{T}\mathbf{\Pi }^{\mathrm{s}}-\mathbf{ad}_{\mathbf{%
V}^{\mathrm{s}}}^{T}(\dot{\mathbf{\Pi }}^{\mathrm{s}}+\mathbf{M}^{\mathrm{s}}%
\dot{\mathbf{V}}^{\mathrm{s}}).  \notag
\end{align}%
Inserting $\mathbf{M}^{\mathrm{s}}\dot{\mathbf{V}}^{\mathrm{s}}=\dot{\mathbf{%
\Pi }}^{\mathrm{s}}+\mathbf{ad}_{\mathbf{V}^{\mathrm{s}}}^{T}\mathbf{\Pi }^{%
\mathrm{s}}$ and rearranging terms yields%
\begin{equation}
\ddot{\mathbf{\Pi }}{^{\mathrm{s}}}=\mathbf{M}^{\mathrm{s}}(\ddot{\mathbf{V}}%
^{\mathrm{s}}-\mathbf{ad}_{\mathbf{V}^{\mathrm{s}}}\dot{\mathbf{V}}^{\mathrm{%
s}})-2\mathbf{ad}_{\mathbf{V}^{\mathrm{s}}}^{T}\dot{\mathbf{\Pi }}^{\mathrm{s%
}}-(\mathbf{ad}_{\dot{\mathbf{V}}^{\mathrm{s}}}+\mathbf{ad}_{\mathbf{V}^{%
\mathrm{s}}}^{2})^{T}\mathbf{\Pi }^{\mathrm{s}}.  \label{Pis2dot}
\end{equation}%
The expression (\ref{Pis2dot}) allows calculating the second time derivative
of the spatial momentum screw from the momentum and its first derivative.

A further time derivative of the NE-equations (\ref{equil1}) determines the
second derivative of the spatial wrench%
\begin{equation}
\dddot{\mathbf{\Pi }}{^{\mathrm{s}}+\ddot{\mathbf{W}}}^{\mathrm{s},\mathrm{%
app}}+\ddot{\mathbf{W}}^{\mathrm{s},\mathrm{grav}}=\mathbf{0}.%
\label{equil3}
\end{equation}%
Starting from (\ref{Pis2dot}) the third derivative of $\mathbf{\Pi }{^{%
\mathrm{s}}}$ is 
\begin{align}
\dddot{\mathbf{\Pi }}{^{\mathrm{s}}}=& \ \mathbf{M}^{\mathrm{s}}(\dddot{%
\mathbf{V}}{^{\mathrm{s}}}-2\mathbf{ad}_{\mathbf{V}^{\mathrm{s}}}\ddot{%
\mathbf{V}}^{\mathrm{s}}+\mathbf{ad}_{\mathbf{V}^{\mathrm{s}}}^{2}\dot{%
\mathbf{V}}^{\mathrm{s}})-2\mathbf{ad}_{\mathbf{V}^{\mathrm{s}}}^{T}\ddot{%
\mathbf{\Pi }}^{\mathrm{s}}  \notag \\
& -\mathbf{ad}_{\mathbf{V}^{\mathrm{s}}}^{T}\mathbf{M}^{\mathrm{s}}(\ddot{%
\mathbf{V}}^{\mathrm{s}}-\mathbf{ad}_{\mathbf{V}^{\mathrm{s}}}\dot{\mathbf{V}%
}^{\mathrm{s}})-(3\mathbf{ad}_{\dot{\mathbf{V}}^{\mathrm{s}}}+\mathbf{ad}_{%
\mathbf{V}^{\mathrm{s}}}^{2})^{T}\dot{\mathbf{\Pi }}^{\mathrm{s}}  \notag \\
& -(\mathbf{ad}_{\ddot{\mathbf{V}}^{\mathrm{s}}}+\mathbf{ad}_{\dot{\mathbf{V}%
}^{\mathrm{s}}}\mathbf{ad}_{\mathbf{V}^{\mathrm{s}}}+\mathbf{ad}_{\mathbf{V}%
^{\mathrm{s}}}\mathbf{ad}_{\dot{\mathbf{V}}^{\mathrm{s}}})^{T}\mathbf{\Pi }^{%
\mathrm{s}}.  \label{Pis3dotTmp}
\end{align}%
Solving (\ref{Pis2dot}) for $\mathbf{M}^{\mathrm{s}}(\ddot{\mathbf{V}}^{%
\mathrm{s}}-\mathbf{ad}_{\mathbf{V}^{\mathrm{s}}}\dot{\mathbf{V}}^{\mathrm{s}%
})$ and replacing this in (\ref{Pis3dotTmp}) leads to the final form of the
third time derivative%
\begin{align}
\dddot{\mathbf{\Pi }}{^{\mathrm{s}}}=& \ \mathbf{M}^{\mathrm{s}}(\dddot{%
\mathbf{V}}{^{\mathrm{s}}}-2\mathbf{ad}_{\mathbf{V}^{\mathrm{s}}}\ddot{%
\mathbf{V}}^{\mathrm{s}}+\mathbf{ad}_{\mathbf{V}^{\mathrm{s}}}^{2}\dot{%
\mathbf{V}}^{\mathrm{s}})  \notag \\
& -3\mathbf{ad}_{\mathbf{V}^{\mathrm{s}}}^{T}\ddot{\mathbf{\Pi }}^{\mathrm{s}%
}-3(\mathbf{ad}_{\dot{\mathbf{V}}^{\mathrm{s}}}+\mathbf{ad}_{\mathbf{V}^{%
\mathrm{s}}}^{2})^{T}\dot{\mathbf{\Pi }}^{\mathrm{s}}  \label{Pis3dot} \\
& -(\mathbf{ad}_{\ddot{\mathbf{V}}^{\mathrm{s}}}+2\mathbf{ad}_{\dot{\mathbf{V%
}}^{\mathrm{s}}}\mathbf{ad}_{\mathbf{V}^{\mathrm{s}}}+\mathbf{ad}_{\mathbf{V}%
^{\mathrm{s}}}\mathbf{ad}_{\dot{\mathbf{V}}^{\mathrm{s}}}+\mathbf{ad}_{%
\mathbf{V}^{\mathrm{s}}}^{3})^{T}\mathbf{\Pi }^{\mathrm{s}}.  \notag
\end{align}

In the expressions (\ref{equil1}), (\ref{Pis2dot}), (\ref{Pis3dot}), the
derivatives of $\mathbf{\Pi }^{\mathrm{s}}$ could be expressed in terms of $%
\mathbf{V}^{\mathrm{s}}$ and its time derivatives. This is not shown since
it is computationally more efficient to reuse the derivatives of $\mathbf{%
\Pi }^{\mathrm{s}}$.

\subsection{End-effector loads and applied wrenches}

The manipulator is subjected to external loads, and in particular interacts
with the environment via the EE. The corresponding EE-wrenches are due to
the (possibly time-varying) inertia of the object being manipulated or due
to contact, for instance. With the assumption that the reference frame $%
\mathcal{F}_{n}$ of the terminal link is located at the EE, the EE-load
contributes an applied wrench $\mathbf{W}_{n}^{\mathrm{s}\text{,}\mathrm{app}%
}\left( t\right) $ at the terminal link. Further general loads acting on the
remaining bodies are included in the model by $\mathbf{W}_{i}^{\mathrm{s}%
\text{,}\mathrm{app}}\left( t\right) ,i=1,\ldots ,n-1$ if they are known.

EE-loads are not included in the recursive higher-order inverse dynamics
algorithms reported in the literature. This is motivated by the fact that
measuring them directly requires force/toque sensors collocated at the EE,
but moreover estimating their time derivatives is difficult \cite{Giusti2018}%
. Nevertheless, for control purposes it is important to include EE-loads,
and their time derivatives whenever possible.

\subsection{Gravity loads}

If the effect of gravity is of interest (e.g. for design purpose), the
gravity wrench and its time derivatives can be determined separately. 
The wrench at body $i$ 
due to gravity in spatial representation is%
\begin{equation}
\mathbf{W}_{i}^{\mathrm{s},\mathrm{grav}}=\mathbf{Ad}_{\mathbf{C}_{i}}^{-T}%
\mathbf{M}_{i}^{\mathrm{b}}\mathbf{Ad}_{\mathbf{C}_{i}}^{-1}\mathbf{G}_{i}^{%
\mathrm{s}}=\mathbf{M}_{i}^{\mathrm{s}}\mathbf{G}_{i}^{\mathrm{s}}.
\label{Wsgrav}
\end{equation}%
Time derivatives of this gravity wrench follow immediately with repeated
application of (\ref{Msdot}) as%
\begin{align}
\dot{\mathbf{W}}_{i}^{\mathrm{s},\mathrm{grav}}& =-(\mathbf{M}_{i}^{\mathrm{s%
}}\mathbf{ad}_{\mathbf{V}_{i}^{\mathrm{s}}}+\mathbf{ad}_{\mathbf{V}_{i}^{%
\mathrm{s}}}^{T}\mathbf{M}_{i}^{\mathrm{s}})\mathbf{G}_{i}^{\mathrm{s}}
\label{Wsgrav1} \\
\ddot{\mathbf{W}}_{i}^{\mathrm{s},\mathrm{grav}}& =%
\Big%
((\mathbf{M}_{i}^{\mathrm{s}}\mathbf{ad}_{\mathbf{V}_{i}^{\mathrm{s}}}^{2}+%
\mathbf{ad}_{\mathbf{V}_{i}^{\mathrm{s}}}^{T}\mathbf{M}_{i}^{\mathrm{s}}%
\mathbf{ad}_{\mathbf{V}_{i}^{\mathrm{s}}}+\mathbf{M}_{i}^{\mathrm{s}}\mathbf{%
ad}_{\dot{\mathbf{V}}_{i}^{\mathrm{s}}})  \label{Wsgrav2} \\
& +(\mathbf{M}_{i}^{\mathrm{s}}\mathbf{ad}_{\mathbf{V}_{i}^{\mathrm{s}}}^{2}+%
\mathbf{ad}_{\mathbf{V}_{i}^{\mathrm{s}}}^{T}\mathbf{M}_{i}^{\mathrm{s}}%
\mathbf{ad}_{\mathbf{V}_{i}^{\mathrm{s}}}+\mathbf{M}_{i}^{\mathrm{s}}\mathbf{%
ad}_{\dot{\mathbf{V}}_{i}^{\mathrm{s}}})^{T}%
\Big%
)\mathbf{G}_{i}^{\mathrm{s}}.  \notag
\end{align}%
The relations (\ref{Wsgrav})-(\ref{Wsgrav2}) can be evaluated within the
forward kinematics recursion (Algorithm 1).%

\subsection{Generalized forces and their derivatives}

The NE-equations (\ref{equil1}) govern the dynamics of the individual
unconstrained bodies. When bodies are regarded as part of the manipulator,
the NE-equations are%
\begin{equation}
\dot{\mathbf{\Pi }}_{i}^{\mathrm{s}}+\mathbf{W}_{i}^{\mathrm{s},\mathrm{app}%
}+\mathbf{W}_{i}^{\mathrm{s},\mathrm{grav}}=\mathbf{W}_{i}^{\mathrm{s}}
\end{equation}%
where $\mathbf{W}_{i}^{\mathrm{s}}$ accounts for constraint reactions as
well as joint forces/torques in the 'free directions'. Applying Jourdain's
principle of virtual power, along with the kinematic relation (\ref{Vsirec}%
), yields%
\begin{align}
\delta \mathbf{V}_{i}^{\mathrm{s}T}\mathbf{W}_{i}^{\mathrm{s}}& =\delta 
\mathbf{V}_{i}^{\mathrm{s}T}(\dot{\mathbf{\Pi }}_{i}^{\mathrm{s}}+\mathbf{W}%
_{i}^{\mathrm{s},\mathrm{app}}+\mathbf{W}_{i}^{\mathrm{s},\mathrm{grav}})
\label{Veriation} \\
\sum_{j=1}^{i}\delta q_{j}\mathbf{S}_{j}^{T}\mathbf{W}_{i}^{\mathrm{s}}&
=\sum_{j=1}^{i}\delta q_{j}\mathbf{S}_{j}^{T}(\dot{\mathbf{\Pi }}_{i}^{%
\mathrm{s}}+\mathbf{W}_{i}^{\mathrm{s},\mathrm{app}}+\mathbf{W}_{i}^{\mathrm{%
s},\mathrm{grav}}).  \notag
\end{align}%
Rearranging relation (\ref{Veriation}) according to $q_{i}$ yields%
\begin{equation}
\delta q_{i}\mathbf{S}_{i}^{T}\sum_{j=i}^{n}\mathbf{W}_{j}^{\mathrm{s}%
}=\delta q_{i}\mathbf{S}_{i}^{T}\sum_{j=i}^{n}(\dot{\mathbf{\Pi }}_{j}^{%
\mathrm{s}}+\mathbf{W}_{j}^{\mathrm{s},\mathrm{app}}+\mathbf{W}_{j}^{\mathrm{%
s},\mathrm{grav}}).
\end{equation}%
The generalized forces $Q_{i}:=\mathbf{S}_{i}^{T}(\mathbf{W}_{i}^{\mathrm{s}%
}+\ldots +\mathbf{W}_{n}^{\mathrm{s}})$ are thus determined by%
\begin{equation}
Q_{i}=\mathbf{S}_{i}^{T}(\dot{\mathbf{\Pi }}_{i}^{\mathrm{s}}+\mathbf{W}%
_{i}^{\mathrm{s},\mathrm{app}}+\mathbf{W}_{i}^{\mathrm{s},\mathrm{grav}}+%
\bar{\mathbf{W}}_{i+1}^{\mathrm{s}})
\end{equation}%
where%
\begin{equation}
\bar{\mathbf{W}}_{i}^{\mathrm{s}}:=\bar{\mathbf{W}}_{i+1}^{\mathrm{s}}+\dot{%
\mathbf{\Pi }}_{i}^{\mathrm{s}}+\mathbf{W}_{i}^{\mathrm{s},\mathrm{app}}+%
\mathbf{W}_{i}^{\mathrm{s},\mathrm{grav}}  \label{Wirec}
\end{equation}%
are the interbody-wrenches due to the (actuated) motion of the kinematic
chain from body $i$ to the terminal body $n$, with $\bar{\mathbf{W}}_{n+1}^{%
\mathrm{s}}=\mathbf{0}$. The term $\mathbf{S}_{i}^{T}\bar{\mathbf{W}}_{i}^{%
\mathrm{s}}$ should be interpreted as reciprocal product of twist (in ray
coordinates) and wrench (in axis coordinates). The time derivatives of the
generalized forces are thus%
\begin{eqnarray}
\dot{Q}_{i} &=&\mathbf{S}_{i}^{T}\dot{\bar{\mathbf{W}}}_{i}^{\text{\textrm{s}%
}}+\dot{\mathbf{S}}_{i}^{T}\bar{\mathbf{W}}_{i}^{\text{\textrm{s}}}
\label{Qi} \\
\ddot{Q}_{i} &=&\mathbf{S}_{i}^{T}\ddot{\bar{\mathbf{W}}}_{i}^{\text{\textrm{%
s}}}+\ddot{\mathbf{S}}_{i}^{T}\bar{\mathbf{W}}_{i}^{\text{\textrm{s}}}+2\dot{%
\mathbf{S}}_{i}^{T}\dot{\bar{\mathbf{W}}}_{i}^{\text{\textrm{s}}}  \notag
\end{eqnarray}

\subsection{Inverse dynamics algorithm}

After execution of the recursive forward kinematics Algorithm 1 (or
Algorithm 2 if the inverse kinematics is not already solved) the
configuration, spatial twist, acceleration, jerk, and jounce of each
individual body are computed. Additionally, the instantaneous joint screws $%
\mathbf{S}_{i}$ and their time derivatives are available. This data provides
the input for the inverse dynamics run. The second-order inverse dynamics
computes the generalized forces $\mathbf{Q}$, i.e. joint forces/torques, and
their first and second time derivatives needed to perform a desired motion $%
\mathbf{q}\left( t\right) $. The relations (\ref{Qi}) and (\ref{Wirec})
together with (\ref{equil1}), (\ref{Pis2dot}), (\ref{Pis3dot}) and (\ref%
{Wsgrav}), (\ref{Wsgrav1}), (\ref{Wsgrav2}) give rise to the backward
recursion listed as Algorithm 3 solving the inverse dynamics problem. The
spatial momentum and its time derivatives are used as intermediate
variables. If available, in addition to the data computed by the forward
kinematics recursion, the applied wrenches, and in particular the wrench due
to EE-loads, is supplied to Algorithm 3. While the momenta $\mathbf{\Pi }%
_{i}^{\mathrm{s}}$ and inter-body wrenches $\bar{\mathbf{W}}_{i}^{\mathrm{s}%
} $, and their time derivatives, are basically algorithmic variables they
can be used to monitor the manipulator.\newline
\begin{figure}[ht]
\begin{tabular}{l}
\textbf{Algorithm 3: 2nd-Order Inverse Dynamics Recursion} \\ 
\hline
\textbf{Input:} $\mathbf{C}_{i},\mathbf{V}_{i}^{\text{\textrm{s}}},\dot{%
\mathbf{V}}_{i}^{\text{\textrm{s}}},\ddot{\mathbf{V}}_{i}^{\text{\textrm{s}}%
},\dddot{\mathbf{V}}{_{i}^{\text{\textrm{s}}},}\mathbf{S}_{i},\dot{\mathbf{S}%
}_{i},\ddot{\mathbf{S}}_{i},\dddot{\mathbf{S}}_{i},i=1,\ldots ,n$ \\ 
$\ \ \ \ \ \ \ \ \ \mathbf{W}_{i}^{\mathrm{s},\mathrm{app}},\dot{\mathbf{W}}%
_{i}^{\text{\textrm{s}},\mathrm{app}},\ddot{\mathbf{W}}_{i}^{\text{\textrm{s}%
},\mathrm{app}},i=1,\ldots ,n;%
\ \dot{\mathbf{V}}_{0}^{\text{\textrm{s}}}%
$ \\ 
For $i=n,\ldots ,1$ do \\ 
\begin{tabular}{rll}
$\mathbf{M}_{i}^{\text{\textrm{s}}}%
\hspace{-2ex}%
$ & $=%
\hspace{-2ex}%
$ & $\mathbf{Ad}_{\mathbf{C}_{i}}^{-T}\mathbf{M}_{i}^{\text{b}}\mathbf{Ad}_{%
\mathbf{C}_{i}}^{-1}$ \\ 
$\mathbf{\Pi }_{i}^{\mathrm{s}}%
\hspace{-2ex}%
$ & $=%
\hspace{-2ex}%
$ & $\mathbf{M}_{i}^{\mathrm{s}}\mathbf{V}_{i}^{\mathrm{s}}$ \\ 
$\dot{\mathbf{\Pi }}{_{i}^{\mathrm{s}}}%
\hspace{-2ex}%
$ & $=%
\hspace{-2ex}%
$ & $\mathbf{M}_{i}^{\mathrm{s}}\dot{\mathbf{V}}_{i}^{\mathrm{s}}-\mathbf{ad}%
_{\mathbf{V}_{i}^{\mathrm{s}}}^{T}\mathbf{\Pi }_{i}^{\mathrm{s}}$ \\ 
$\ddot{\mathbf{\Pi }}{_{i}^{\mathrm{s}}}%
\hspace{-2ex}%
$ & $=%
\hspace{-2ex}%
$ & $\mathbf{M}_{i}^{\mathrm{s}}(\ddot{\mathbf{V}}_{i}^{\mathrm{s}}-\mathbf{%
ad}_{\mathbf{V}_{i}^{\mathrm{s}}}\dot{\mathbf{V}}_{i}^{\mathrm{s}})$ \\ 
&  & $-2\mathbf{ad}_{\mathbf{V}_{i}^{\mathrm{s}}}^{T}\dot{\mathbf{\Pi }}%
_{i}^{\mathrm{s}}-(\mathbf{ad}_{\dot{\mathbf{V}}_{i}^{\mathrm{s}}}+\mathbf{ad%
}_{\mathbf{V}_{i}^{\mathrm{s}}}^{2})^{T}\mathbf{\Pi }_{i}^{\mathrm{s}}$ \\ 
$\dddot{\mathbf{\Pi }}{_{i}^{\mathrm{s}}}%
\hspace{-2ex}%
$ & $=%
\hspace{-2ex}%
$ & $\mathbf{M}_{i}^{\mathrm{s}}(\dddot{\mathbf{V}}{_{i}^{\mathrm{s}}}-2%
\mathbf{ad}_{\mathbf{V}_{i}^{\mathrm{s}}}\ddot{\mathbf{V}}_{i}^{\mathrm{s}}+%
\mathbf{ad}_{\mathbf{V}_{i}^{\mathrm{s}}}^{2}\dot{\mathbf{V}}_{i}^{\mathrm{s}%
})$ \\ 
&  & $-3\mathbf{ad}_{\mathbf{V}_{i}^{\mathrm{s}}}^{T}\ddot{\mathbf{\Pi }}%
_{i}^{\mathrm{s}}-3(\mathbf{ad}_{\dot{\mathbf{V}}_{i}^{\mathrm{s}}}+\mathbf{%
ad}_{\mathbf{V}_{i}^{\mathrm{s}}}^{2})^{T}\dot{\mathbf{\Pi }}_{i}^{\mathrm{s}%
}$ \\ 
&  & $-(\mathbf{ad}_{\ddot{\mathbf{V}}_{i}^{\mathrm{s}}}+2\mathbf{ad}_{\dot{%
\mathbf{V}}_{i}^{\mathrm{s}}}\mathbf{ad}_{\mathbf{V}_{i}^{\mathrm{s}}}+%
\mathbf{ad}_{\mathbf{V}_{i}^{\mathrm{s}}}\mathbf{ad}_{\dot{\mathbf{V}}_{i}^{%
\mathrm{s}}}+\mathbf{ad}_{\mathbf{V}_{i}^{\mathrm{s}}}^{3})^{T}\mathbf{\Pi }%
_{i}^{\mathrm{s}}$ \\ 
$\bar{\mathbf{W}}_{i}^{\text{\textrm{s}}}%
\hspace{-2ex}%
$ & $=%
\hspace{-2ex}%
$ & $%
\bar{\mathbf{W}}_{i+1}^{\text{\textrm{s}}}+\dot{\mathbf{\Pi }}_{i}^{\mathrm{s%
}}+\bar{\mathbf{W}}_{i}^{\mathrm{s},\mathrm{app}}%
$ \\ 
$\dot{\bar{\mathbf{W}}}_{i}^{\text{\textrm{s}}}%
\hspace{-2ex}%
$ & $=%
\hspace{-2ex}%
$ & $%
\dot{\bar{\mathbf{W}}}_{i+1}^{\text{\textrm{s}}}+\ddot{\mathbf{\Pi }}{_{i}^{%
\mathrm{s}}}+\dot{\bar{\mathbf{W}}}_{i}^{\text{\textrm{s}},\mathrm{app}}%
$ \\ 
$\ddot{\bar{\mathbf{W}}}_{i}^{\text{\textrm{s}}}%
\hspace{-2ex}%
$ & $=%
\hspace{-2ex}%
$ & $%
\ddot{\bar{\mathbf{W}}}_{i+1}^{\text{\textrm{s}}}+\dddot{\mathbf{\Pi }}{%
_{i}^{\mathrm{s}}}+\ddot{\bar{\mathbf{W}}}_{i}^{\text{\textrm{s}},\mathrm{app%
}}%
$%
\vspace{0.5ex}
\\ 
$Q_{i}%
\hspace{-2ex}%
$ & $=%
\hspace{-2ex}%
$ & $\mathbf{S}_{i}^{T}\bar{\mathbf{W}}_{i}^{\text{\textrm{s}}}$ \\ 
$\dot{Q}_{i}%
\hspace{-2ex}%
$ & $=%
\hspace{-2ex}%
$ & $\mathbf{S}_{i}^{T}\dot{\bar{\mathbf{W}}}_{i}^{\text{\textrm{s}}}+\dot{%
\mathbf{S}}_{i}^{T}\bar{\mathbf{W}}_{i}^{\text{\textrm{s}}}$ \\ 
$\ddot{Q}_{i}%
\hspace{-2ex}%
$ & $=%
\hspace{-2ex}%
$ & $\mathbf{S}_{i}^{T}\ddot{\bar{\mathbf{W}}}_{i}^{\text{\textrm{s}}}+\ddot{%
\mathbf{S}}_{i}^{T}\bar{\mathbf{W}}_{i}^{\text{\textrm{s}}}+2\dot{\mathbf{S}}%
_{i}^{T}\dot{\bar{\mathbf{W}}}_{i}^{\text{\textrm{s}}}$%
\end{tabular}
\\ 
End \\ 
\textbf{Output:} $Q_{i},%
\dot{Q}_{i},\ddot{Q}_{i}%
$ and inter-body wrenches $\bar{\mathbf{W}}_{i}^{\text{\textrm{s}}},\dot{%
\bar{\mathbf{W}}}_{i}^{\text{\textrm{s}}},\ddot{\bar{\mathbf{W}}}_{i}^{\text{%
\textrm{s}}}$ \\ \hline
\end{tabular}
\end{figure}

\subsection{Features of the algorithm}

The basic difference of the spatial formulation, in contrast to the
body-fixed formulation (see Appendix \ref{secBodyFixedAlg}), is that the
recursions (\ref{Vsirec}) and (\ref{Wirec}) do not involve frame
transformations of twists or wrenches. This is obvious comparing (\ref%
{Vsirec}) with (\ref{Vbrec}), and (\ref{Wirec}) with (\ref{Wbrec}). However,
in the spatial formulation (\ref{Vsirec}), the instantaneous joint screw
coordinates ${\mathbf{S}}_{i}$ must be computed by a frame transformation of 
${\mathbf{Y}}_{i}$, while the ${^{i}\mathbf{X}}_{i}$ in (\ref{Vbrec}) are
constant. Further, in (\ref{NESpatial}) the mass matrix is not constant and
must be computed according to (\ref{Ms}). Therefore the spatial and
body-fixed versions are computationally equivalent. A detailed analysis can
be found in \cite{Featherstone2008} where it is shown that the body-fixed
formulation requires slightly less numerical operations. This argument does
not carry over to the higher-order algorithms, however, since the joint
screws ${\mathbf{S}}_{i}$ and the mass matrices $\mathbf{M}_{i}^{\mathrm{s}}$
are reused in the higher-order relations of Algorithm 1 and 2, and no frame
transformations of the time derivatives of $\mathbf{V}_{i}^{\mathrm{s}}$ and 
$\bar{\mathbf{W}}_{i}^{\mathrm{s}}$ are necessary. Compared with the
algorithm in body-fixed representation (Appendix \ref{secBodyFixedAlg}),
computing the time derivatives $\dot{Q}_{i},\ddot{Q}_{i}$ is slightly more
expensive as it involves two, respectively three scalar products. Also the
time derivatives of the NE-equations involve an additional term.

The higher-order algorithm inherits the basic features of the Lie group
formulations for robot kinematics and dynamics \cite%
{LynchPark2017,MUBOScrews1,MUBOScrews2}. The most advantageous feature of
this geometric formulation is the frame-invariant parameterization in terms
of vectorial quantities. 
This allows, for instance, processing data in universal robotic description
format (URDF), which describe joint and link kinematics in terms of position
vectors and rotation matrices \cite{ROS-URDF,LynchPark2017}.%

\section{Example: Franka Emika Panda%
\label{secIPanda}%
}

The Franka Emika Panda is a 
redundant 
7 DOF robotic arm. Its dynamic parameters were reported in \cite{Gaz2019}.
In the zero reference configuration, shown in Fig. 1 of \cite{Gaz2019}, the
unit vectors along the joint axes, resolved in the IFR $\mathcal{F}_{0}$, are%
\begin{align*}
\mathbf{e}_{1}& =\mathbf{e}_{3}=\mathbf{e}_{5}=\left( 0,0,1\right) ^{T},%
\mathbf{e}_{2}=\left( 0,1,0\right) ^{T} \\
\mathbf{e}_{4}& =\mathbf{e}_{6}=\left( 0,-1,0\right) ^{T},\mathbf{e}%
_{7}=\left( 0,0,-1\right) ^{T}
\end{align*}%
and the position vectors to the joint axes are%
\begin{align*}
\mathbf{y}_{1}& =\left( 0,0,d_{1}\right) ^{T},\mathbf{y}_{3}=\left(
0,0,d_{1}+d_{3}\right) ^{T},\mathbf{y}_{4}=\left( a_{4},0,d_{1}+d_{3}\right)
^{T} \\
\mathbf{y}_{5}& =\left( 0,0,d_{1}+d_{3}+d_{5}\right) ^{T},\mathbf{y}%
_{7}=\left( a_{7},0,d_{1}+d_{3}+d_{5}\right) ^{T},
\end{align*}%
and $\mathbf{y}_{2}=\mathbf{y}_{1},\mathbf{y}_{6}=\mathbf{y}_{5}$. The
geometric parameters given in the data sheet are $d_{1}=0.333$m$,d_{3}=0.316$%
m$,d_{5}=0.384$m, $a_{4}=0.0825$m$,a_{7}=0.088$m. The spatial screw
coordinates according to $\mathbf{Y}_{j}=\left( \mathbf{e}_{j},\mathbf{y}%
_{j}\right) ^{T}$ are thus%
\begin{align*}
\mathbf{Y}_{1}& =\mathbf{Y}_{3}=\mathbf{Y}_{5}=\left( 0,0,1,0,0,0\right) ^{T}
\\
\mathbf{Y}_{2}& =\left( 0,1,0,-d_{1},0,0\right) ^{T},\mathbf{Y}_{4}=\left(
0,-1,0,d_{1}+d_{3},0,-a_{4}\right) ^{T} \\
\mathbf{Y}_{6}& =\left( 0,-1,0,d_{1}+d_{3}+d_{5},0,0\right) ^{T},\mathbf{Y}%
_{7}=\left( 0,0,-1,0,a_{7},0\right) ^{T}.
\end{align*}%
Body-fixed reference frames are defined as indicated in Fig. 1 of \cite%
{Gaz2019}. They are located at the joints so that the $\mathbf{y}_{i}$ are
also position vectors to the reference frames. The reference configuration
of the bodies are 
\begin{align*}
\mathbf{A}_{1}& =\left( {\scriptsize 
\begin{array}{cccc}
1 & 0 & 0 & 0 \\ 
0 & 1 & 0 & 0 \\ 
0 & 0 & 1 & 0.333 \\ 
0 & 0 & 0 & 1%
\end{array}%
}\right) ,\mathbf{A}_{2}=\left( {\scriptsize 
\begin{array}{cccc}
1 & 0 & 0 & 0 \\ 
0 & 0 & 1 & 0 \\ 
0 & -1 & 0 & 0.333 \\ 
0 & 0 & 0 & 1%
\end{array}%
}\right) \\
\mathbf{A}_{3}& =\left( {\scriptsize 
\begin{array}{cccc}
1 & 0 & 0 & 0 \\ 
0 & 1 & 0 & 0 \\ 
0 & 0 & 1 & 0.649 \\ 
0 & 0 & 0 & 1%
\end{array}%
}\right) ,\mathbf{A}_{4}=\left( {\scriptsize 
\begin{array}{cccc}
1 & 0 & 0 & 0.0825 \\ 
0 & 0 & -1 & 0 \\ 
0 & 1 & 0 & 0.649 \\ 
0 & 0 & 0 & 1%
\end{array}%
}\right) \\
\mathbf{A}_{5}& =\left( {\scriptsize 
\begin{array}{cccc}
1 & 0 & 0 & 0 \\ 
0 & 1 & 0 & 0 \\ 
0 & 0 & 1 & 1.033 \\ 
0 & 0 & 0 & 1%
\end{array}%
}\right) ,\mathbf{A}_{6}=\left( {\scriptsize 
\begin{array}{cccc}
1 & 0 & 0 & 0 \\ 
0 & 0 & -1 & 0 \\ 
0 & 1 & 0 & 1.033 \\ 
0 & 0 & 0 & 1%
\end{array}%
}\right)
\end{align*}%
\begin{equation*}
\mathbf{A}_{7}=\left( {\scriptsize 
\begin{array}{cccc}
1 & 0 & 0 & 0.088 \\ 
0 & -1 & 0 & 0 \\ 
0 & 0 & -1 & 1.033 \\ 
0 & 0 & 0 & 1%
\end{array}%
}\right)
\end{equation*}%
The dynamic parameters (mass, COM, inertia) w.r.t. to the body-fixed frames
are taken from Tables VIII and IX of the supplementary document to \cite%
{Gaz2019}.

\begin{figure}[tb]
\begin{center}
\includegraphics[width=8.0cm]{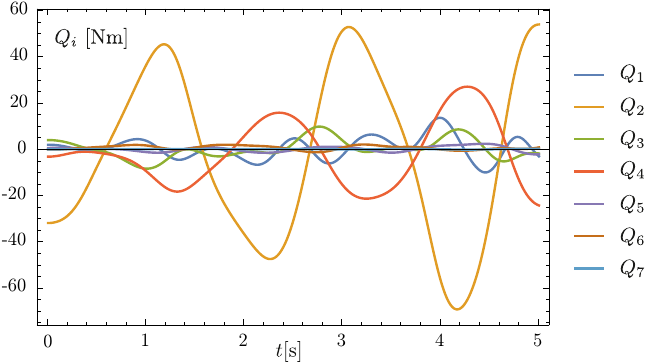} %
\includegraphics[width=8cm]{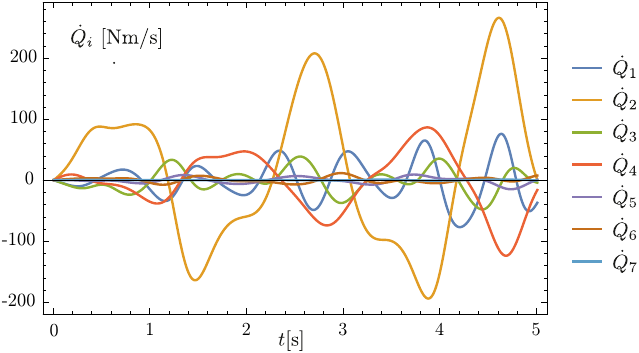} %
\includegraphics[width=8cm]{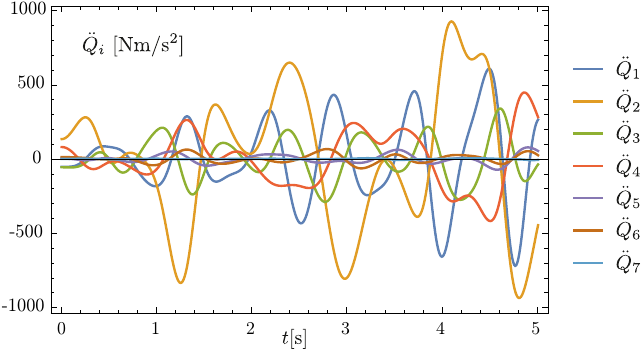}
\end{center}
\caption{Joint torques $Q_{i}$ and their first and second derivatives, $\dot{%
Q}_{i}$ and $\ddot{Q}_{i}$, for the validation trajectory according to
equation (31) in \protect\cite{Gaz2019}.}
\label{figQ}
\end{figure}

The spatial forward kinematics algorithm 1 and the inverse dynamics
algorithm 3 have been implemented in Matlab. Also implemented was the
body-fixed version of the recursive algorithm from \cite{ICRA2017}. The code
and example scripts are available as supplementary Multimedia Material. The
second-order inverse dynamics was solved for the validation trajectory $%
\mathbf{q}\left( t\right) $ defined by (31) in \cite{Gaz2019}. This allows
direct comparison with results reported in \cite{Gaz2019}, which can be
reproduced using the toolbox referred to in \cite{Gaz2019}. 
The inverse kinematics algorithm 2 could be used if the EE motion is given,
rather than the joint trajectory. To this end, one could resort to the joint
space decomposition since the Panda robot is kinematically redundant. 
Note that friction was not included for all computations. Figure \ref{figQ}
shows the joint torques $Q_{i}$ and their derivatives $\dot{Q}_{i},\ddot{Q}%
_{i}$. 
The results were validated against the solution computed from the closed
form analytic expressions of the EOM and their analytic derivatives.

As a preliminary comparison of the computational performance, the execution
times for 40000 evaluation calls of the Matlab code for the presented
spatial formulation and for the body-fixed algorithm \cite{ICRA2017} were
determined. On a PC (i5-9500, 3GHz) running MS-Windows, the spatial
algorithm needed 7.15\thinspace s and the body-fixed 8.36\thinspace s, i.e.
the spatial is 15\% faster. These preliminary results are not
representative, however, since the timing includes computational overhead
and also Matlab's run time optimization. A reliable comparison would
necessitate a dedicated stand-alone implementation, preferably on a
real-time system.%

\section{Conclusion and suggested Future Research%
\label{secConclusion}%
}

The proposed recursive formulation for the second-order inverse dynamics and
the fourth-order forward kinematics allow for implementation of
computationally efficient $O\left( n\right) $ algorithms. The presented
algorithm in spatial representation complements the Lie group formulation in
body-fixed and hybrid representation reported in \cite{ICRA2017}. Therewith,
second order inverse-dynamics $O\left( n\right) $ algorithms are now
available for the three relevant representations of spatial objects. This
provides the basis for an exhaustive comparison and investigation into their
computational efficiency \emph{and} their user-friendliness. Moreover, in
contrast to the formulations using classical vector algebra, the Lie group
formulation provides an alternative that is compact, and easy to maintain
and implement, which parallels the EOM Lie group formulations \cite%
{LynchPark2017,MUBOScrews2}. 
When rolled out and written in terms of the 3-vector notation, the
body-fixed formulation \cite{ICRA2017} gives exactly the relations presented
in \cite{BuondonnaDeLuca2015,BuondonnaDeLuca2016,Guarino2006,Guarino2009}.
It should also be remarked that the actual representation (spatial,
body-fixed, hybrid) of spatial objects (twists, wrenches, inertias) is
rather an algorithmic feature but is not apparent to the user. The results
for $Q_{i}$ and their derivatives are indeed independent from the
representation.%

Future work will be focused on detailed analysis and comparison of the
complexity of the respective formulation. Special emphasis will be given on
computationally efficient implementations making repeated use of terms as
well as minimizing matrix-vector operations. Only then a final conclusion as
to which representation leads to the most efficient algorithm can be drawn. 
An important aspect of the implementation will be to reduce accumulation of
errors due to the computation with finite precision. The latter could be an
issue in particular for the higher-order inverse dynamics since the
numerical values of $Q_{i},\dot{Q}_{i}$, and $\ddot{Q}_{i}$ have a different
order of magnitude. Another topic will be the higher-order partial
derivatives of the EOM w.r.t. generalized coordinates $q_{i}$. Such
first-order derivatives (using body-fixed representation) were presented in 
\cite{CarpentierMansardRSS 2018,Carpentier-SICE2019}. All formulations
should be compared with automatic differentiation methods \cite%
{Giftthaler-AR2017}.%

\appendices%

\section{Screws and Rigid Body Motions}

In the following, basic concepts are recalled as far as necessary for the
paper. Details can be found in \cite{LynchPark2017,Murray,Selig}.

Rigid body configurations are represented by homogenous transformation
matrices that form the Lie group $SE\left( 3\right) $. The configuration of
body $i$, i.e. the configuration of its body-fixed frame $\mathcal{F}_{i}$
w.r.t., the IFR, is then denoted by the matrix 
\begin{equation}
\mathbf{C}_{i}=\left( 
\begin{array}{cc}
\mathbf{R}_{i} & \mathbf{r}_{i} \\ 
\mathbf{0}\ \  & 1%
\end{array}%
\right) \in SE\left( 3\right)  \label{Ci}
\end{equation}%
with rotation matrix 
\mbox{$\mathbf{R}_{i}\in SO\left( 3\right) $ and position
vector $\mathbf{r}_{i}\in {\mathbb{R}}^{3}$}. The time evolution of $\mathbf{%
C}_{i}\left( t\right) $ describes the spatial motion.

Rigid body twists and wrenches are coordinate invariant objects. Twists are
screws that form the Lie algebra $se\left( 3\right) $ and wrenches are
co-screws forming the dual $se^{\ast }\left( 3\right) $. In a chosen
reference frame, both are represented as 6-vectors. Twists $\mathbf{V}%
=\left( \bm{\omega},\mathbf{v}\right) ^{T}$ are represented in ray
coordinates, and wrenches $\mathbf{W}=\left( \mathbf{m},\mathbf{f}\right)
^{T}$ in axis coordinates. Denote with ${^{i}}\mathbf{X}\in {\mathbb{R}}^{6}$
the screw coordinate vector represented in body-fixed frame $\mathcal{F}_{i}$
and with $\mathbf{X}\in {\mathbb{R}}^{6}$ its representation in the spatial
IFR. Screw coordinate vectors transform from body-fixed to spatial
representation according to $\mathbf{X}=\mathbf{Ad}_{\mathbf{C}_{i}}{^{i}}%
\mathbf{X}$, with%
\begin{equation}
\mathbf{Ad}_{\mathbf{C}_{i}}=\left( 
\begin{array}{cc}
\mathbf{R}_{i} & \mathbf{0} \\ 
\widetilde{\mathbf{r}}_{i}\mathbf{R}_{i}\ \  & \mathbf{R}_{i}%
\end{array}%
\right) .  \label{Ad}
\end{equation}%
The corresponding transformation of co-screws (wrenches) is ${^{i}}\mathbf{W}%
=\mathbf{Ad}_{\mathbf{C}_{i}}^{T}\mathbf{W}$, where ${^{i}}\mathbf{W},%
\mathbf{W}\in se^{\ast }\left( 3\right) $ is the co-screw coordinate vector
in body-fixed and spatial representation, respectively. The time derivative
of screw and co-screw coordinates represented in moving frame $\mathcal{F}%
_{i}$ is, respectively,%
\begin{align}
\mathbf{Ad}_{\mathbf{C}_{i}}^{-1}\dot{\mathbf{X}}& ={^{i}}\dot{\mathbf{X}}+%
\mathbf{ad}_{\mathbf{V}_{i}^{\text{b}}}{^{i}\mathbf{X}},\ \text{for }\mathbf{%
X}\in se\left( 3\right)  \label{Vbdot} \\
\mathbf{Ad}_{\mathbf{C}_{i}}^{T}\dot{\mathbf{W}}& ={^{i}}\dot{\mathbf{W}}-%
\mathbf{ad}_{\mathbf{V}_{i}^{\text{b}}}^{T}{^{i}\mathbf{W}},\ \text{for }%
\mathbf{W}\in se^{\ast }\left( 3\right) 
\label{Wbdot}
\end{align}

Denote with $\mathbf{X}=\left( \bm{\xi},\bm{\eta}\right) ^{T}$ a
general screw coordinate vector. The Lie bracket (screw product) is given as 
\begin{equation}
\lbrack \mathbf{X}_{1}{,\mathbf{X}}_{2}]=\left( \bm{\xi}_{1}\times 
\bm{\xi}_{2},\bm{\eta}_{1}\times \bm{\xi}_{2}+\bm{\xi}%
_{1}\times \bm{\eta}_{2}\right) ^{T}=\mathbf{ad}_{\mathbf{X}_{1}}%
\mathbf{X}_{2}  \label{ScrewProd}
\end{equation}%
with the matrix%
\vspace{-2ex}%
\begin{equation}
\mathbf{ad}_{\mathbf{X}}=\left( 
\begin{array}{cc}
\widetilde{\bm{\xi}}\ \  & \mathbf{0} \\ 
\widetilde{\bm{\eta}}\ \  & \widetilde{\bm{\xi}}%
\end{array}%
\right) .  \label{adse3}
\end{equation}%
This operation is occasionally called the 'generalized cross product' \cite%
{Featherstone2008}. The time derivative of the adjoint transformation matrix
and its inverse is%
\begin{eqnarray}
\dot{\mathbf{Ad}}_{\mathbf{C}} &=&%
\mathbf{ad}_{\mathbf{V}^{\mathsf{s}}}\mathbf{Ad}_{\mathbf{C}}%
\label{Addot} \\
\dot{\mathbf{Ad}}_{\mathbf{C}}^{-1} &=&-\mathbf{Ad}_{\mathbf{C}}^{-1}\mathbf{%
ad}_{\mathbf{V}^{\mathsf{s}}}  \label{AdInvdot}
\end{eqnarray}

\section{Main Steps of the Body-Fixed $O\left( n\right) $ Algorithm%
\label{secBodyFixedAlg}%
}

Here some basic recursive relations of the body-fixed algorithm reported in 
\cite{ICRA2017} are shown for comparison. 
The twist of body $i$ in body-fixed representation is the aggregate $\mathbf{%
V}_{i}^{\text{b}}=(\bm{\omega}_{i}^{\text{b}},\mathbf{v}_{i}^{\text{b}%
})^{T}$ of the angular velocity $\bm{\omega}_{i}^{\text{b}}$ of a
body-fixed frame $\mathcal{F}_{i}$ relative to a world frame $\mathcal{F}%
_{0} $, and the translational velocity $\mathbf{v}^{\text{b}}$ of the origin
of $\mathcal{F}_{i}$ relative to $\mathcal{F}_{0}$, where both vectors are
resolved in $\mathcal{F}_{i}$. It is related to the spatial representation
by $\mathbf{V}_{i}^{\text{s}}=\mathbf{Ad}_{\mathbf{C}_{i}}\mathbf{V}_{i}^{%
\text{b}}$. 
The velocities and accelerations are determined in the forward kinematics
run as (${^{i}\mathbf{X}}_{i}$ are constant screw coordinates of joint $i$
represented in the body-fixed frame)%
\begin{eqnarray}
\mathbf{V}_{i}^{\text{b}} &=&\mathbf{Ad}_{\mathbf{C}_{i,i-1}}\mathbf{V}%
_{i-1}^{\text{b}}+{^{i}\mathbf{X}}_{i}\dot{q}_{i}  \label{Vbrec} \\
\dot{\mathbf{V}}_{i}^{\text{b}} &=&\mathbf{Ad}_{\mathbf{C}_{i,i-1}}\dot{%
\mathbf{V}}_{i-1}^{\text{b}}+\dot{q}_{i}\mathbf{ad}_{\mathbf{V}_{i}^{\text{b}%
}}{^{i}\mathbf{X}}_{i}+{^{i}\mathbf{X}}_{i}\ddot{q}_{i}.  \notag
\end{eqnarray}%
The jerk and jounce recursions are omitted. The inter-body wrenches and
their first time derivatives are calculated with%
\begin{eqnarray}
\bar{\mathbf{W}}_{i}^{\text{b}}%
\hspace{-1.6ex}
&=&%
\hspace{-1.6ex}%
\mathbf{Ad}_{\mathbf{C}_{i+1,i}}^{T}\bar{\mathbf{W}}_{i+1}^{\text{b}}+%
\mathbf{M}_{i}^{\text{b}}\dot{\mathbf{V}}_{i}^{\text{b}}-\mathbf{ad}_{%
\mathbf{V}_{i}^{\text{b}}}^{T}\mathbf{M}_{i}^{\text{b}}\mathbf{V}_{i}^{\text{%
b}}+\mathbf{W}_{i}^{\text{b,app}}  \notag \\
\dot{\bar{\mathbf{W}}}_{i}^{\text{b}}%
\hspace{-1.6ex}
&=&%
\hspace{-1.6ex}%
\mathbf{Ad}_{\mathbf{C}_{i+1,i}}^{T}(\dot{\bar{\mathbf{W}}}_{i+1}^{\text{b}}-%
\dot{q}_{i+1}\mathbf{ad}_{{^{i+1}\mathbf{X}}_{i+1}}^{T}\bar{\mathbf{W}}%
_{i+1}^{\text{b}})  \label{Wbrec} \\
&&%
\hspace{-1.6ex}%
+\,\mathbf{M}_{i}^{\text{b}}\ddot{\mathbf{V}}_{i}^{\text{b}}-\mathbf{ad}_{%
\mathbf{V}_{i}^{\text{b}}}^{T}\mathbf{M}_{i}^{\text{b}}\dot{\mathbf{V}}_{i}^{%
\text{b}}-\mathbf{ad}_{\dot{\mathbf{V}}_{i}^{\text{b}}}^{T}\mathbf{M}_{i}^{%
\text{b}}\mathbf{V}_{i}^{\text{b}}+\dot{\mathbf{W}}_{i}^{\text{b,app}} 
\notag \\
Q_{i}%
\hspace{-1.6ex}
&=&%
\hspace{-1.6ex}%
{^{i}\mathbf{X}}_{i}^{T}\bar{\mathbf{W}}_{i}^{\text{b}},\ \ \dot{Q}_{i}={^{i}%
\mathbf{X}}_{i}^{T}\dot{\bar{\mathbf{W}}}_{i}^{\text{b}}.
\end{eqnarray}


\end{document}